\documentclass[11pt]{article}
\usepackage[preprint]{acl}
\usepackage{times}
\usepackage{latexsym}
\usepackage[T1]{fontenc}
\usepackage[utf8]{inputenc}
\usepackage{microtype}
\usepackage{inconsolata}
\usepackage{graphicx}
\usepackage{amsmath,amssymb}
\usepackage{marvosym}
\usepackage{booktabs,tabularx,array}
\usepackage{multirow}
\usepackage{listings}
\usepackage{algorithm}
\usepackage{algpseudocode}
\usepackage{placeins}
\usepackage[skins,breakable,listings]{tcolorbox}
\usepackage{xcolor}
\usepackage{colortbl}
\definecolor{TableHeader}{HTML}{E8EEF7}
\definecolor{TableStripe}{HTML}{F5F7FA}
\definecolor{TableModel}{HTML}{F1EDFC}
\definecolor{TableHighlight}{HTML}{E6EDF8}
\definecolor{TableInk}{HTML}{293B54}
\definecolor{PromptBackground}{HTML}{F8FAFD}
\definecolor{PromptBorder}{HTML}{8FA6BF}
\definecolor{PromptHeader}{HTML}{E8EEF7}
\definecolor{PromptInk}{HTML}{293B54}
\tcbset{paperprompt/.style={
  enhanced jigsaw,breakable,listing only,
  colback=PromptBackground,colframe=PromptBorder,
  colbacktitle=PromptHeader,coltitle=PromptInk,
  fonttitle=\sffamily\bfseries\small,
  boxrule=0.5pt,leftrule=2pt,arc=2pt,
  left=6pt,right=6pt,top=5pt,bottom=5pt,
  toptitle=4pt,bottomtitle=4pt,
  before skip=8pt,after skip=8pt,
  lines before break=4,
  listing options={basicstyle=\ttfamily\small,
    breaklines=true,breakatwhitespace=false,breakindent=0pt,
    columns=fullflexible,keepspaces=true,showstringspaces=false,
    aboveskip=0pt,belowskip=0pt,tabsize=2}
}}

\newcommand{\promptfile}[2]{\par\begin{nolinenumbers}%
  \tcbinputlisting{paperprompt,title={#1},
    title after break={#1\space (continued)},listing file={#2}}%
  \end{nolinenumbers}\par}
\algrenewcommand\algorithmicrequire{\textbf{Input:}}
\algrenewcommand\algorithmicensure{\textbf{Output:}}
\algrenewcommand\alglinenumber[1]{\scriptsize\color{TableInk}#1:}
\algrenewcommand\algorithmicindent{1em}
\newcommand{\personamodel}{\textbf{PersonaAgent:}\enspace
  \raisebox{-0.2ex}{\includegraphics[height=1.4ex]{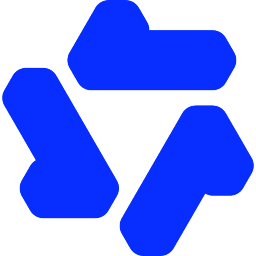}}\enspace Qwen3-8B}
\newcommand{\pibench}{$\pi$-Bench}
\newcommand{\code}[1]{\texttt{#1}}
\newcolumntype{Y}{>{\raggedright\arraybackslash}X}
\newcolumntype{Z}{>{\centering\arraybackslash}X}
\title{AutoPersona: Continuous Personalization Injection\\for Long-Horizon Tasks}
\author{Qijia Zhuang\textsuperscript{1,*}\quad
  Zihan Shen\textsuperscript{1,*}\quad
  Rui Liu\textsuperscript{2}\quad
  Yuxiang Ren\textsuperscript{1,\Letter}\\[0.3em]
  \normalfont\textsuperscript{1}\,Nanjing University\quad
  \textsuperscript{2}\,Northeastern University}
\hypersetup{pdftitle={AutoPersona: Continuous Personalization Injection for Long-Horizon Tasks},
  pdfauthor={Qijia Zhuang, Zihan Shen, Rui Liu, Yuxiang Ren}}
\begin{document}
\maketitle
\begin{abstract}
Personal assistant agents increasingly perform long-horizon tasks that require planning and tool use across multiple stages. Personalization is essential for understanding user needs and providing continued assistance. Existing memory systems primarily support the storage, retrieval, and updating of historical information. However, supplying historical context alone does not ensure that relevant preferences remain effective throughout a task chain, while processing memory sequentially before each subtask adds latency. We introduce AutoPersona, a continuous personalization injection framework for long-horizon tasks. AutoPersona organizes task state and user preferences into Workspace Memory and Persona Memory, and separates personalized memory use from general task execution. A main agent plans and executes an ordered task chain, while a dedicated PersonaAgent retrieves relevant memories and converts them into personalized conditions and guidance injected before each subtask. Supervised fine-tuning and reinforcement learning train PersonaAgent to select and summarize task-relevant information. An asynchronous collaboration mechanism prepares guidance for subsequent subtasks while the main agent executes the current one, overlapping memory processing with task execution. Experiments on VitaBench~2.0, \pibench{}, and Persona2Web show improvements in personalized task accuracy and execution speed.
\end{abstract}

\section{Introduction}
Long-horizon personal assistant tasks involve several stages, such as reading materials, selecting an option, operating tools, analyzing results, and producing a deliverable. Agents must use previous results and workspace state while allowing user preferences to shape subsequent decisions \citep{ref24,ref25,ref41}. Long-term memory benchmarks examine information extraction across sessions, temporal reasoning, and knowledge updating \citep{ref19,ref32}. MemoryArena further couples memory with action in interdependent multi-session tasks \citep{ref08}. These settings motivate a central question: how can historical information become personalized guidance for different execution stages without introducing excessive latency?

As a task progresses, the user preferences and constraints relevant to each stage change. Historical context supplied only at initialization may not remain effective through subsequent reasoning and tool use: preferences may be overlooked, while new results or preference updates can make the initial context outdated. Long-horizon personalization therefore requires relevant preferences to keep influencing later decisions as execution progresses.

Increasing memory-use frequency also incurs costs. Performing retrieval, selection, and information organization sequentially at every stage adds waiting time. Repeatedly supplying extensive histories increases input tokens and introduces information unrelated to the current goal. Long-horizon personalization must maintain the effective use of important preferences while controlling the waiting and context costs of repeated memory processing.

\begin{figure}[t]
\centering
\includegraphics[width=\columnwidth]{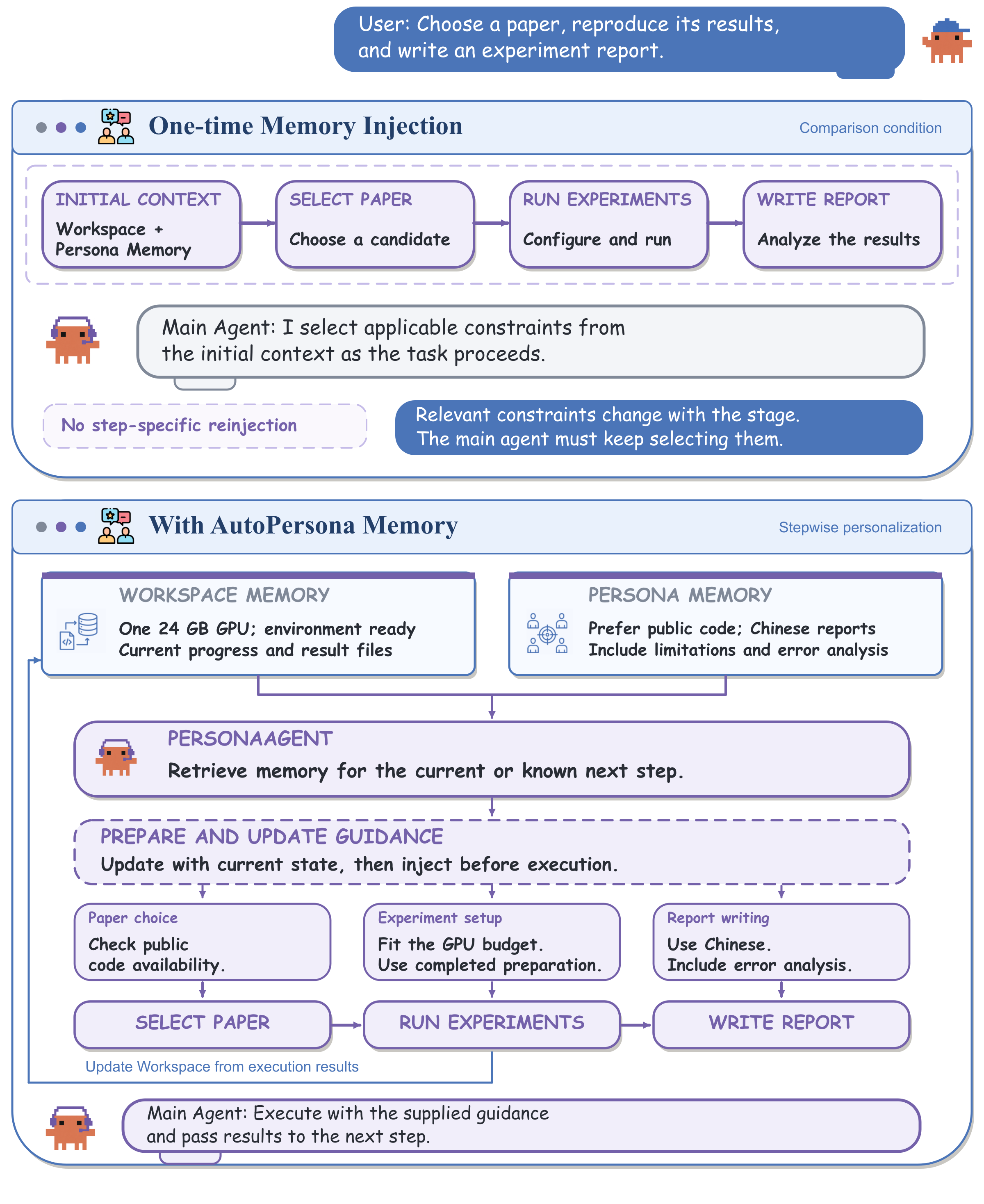}
\caption{One-time memory injection and continuous subtask-level injection in the same paper-reproduction task. The upper workflow supplies memory at task initialization. The lower workflow uses PersonaAgent to prepare guidance, update it with new results, and inject it before each subtask. Execution results update Workspace Memory.}
\label{fig:motivation}
\end{figure}

Figure~\ref{fig:motivation} uses a paper-reproduction task to illustrate one-time and continuous injection, showing a setting in which personalization must span multiple execution stages.

We propose AutoPersona, a continuous personalization injection framework for long-horizon tasks. The main agent plans and advances the task chain. A dedicated PersonaAgent retrieves Workspace and Persona memories and generates subtask-specific conditions and guidance, which are updated with task results and injected before the next subtask. Supervised fine-tuning (SFT) learns retrieval actions and structured guidance; reinforcement learning (RL) optimizes the selection of necessary memory conditions. Asynchronous collaboration overlaps next-subtask preparation with current execution.

We evaluate AutoPersona on VitaBench~2.0, \pibench{}, and Persona2Web \citep{ref42,ref37,ref14}. The results show improved accuracy and execution speed in personalized tasks. Our contributions are:
\begin{itemize}
\item A continuous personalization injection framework that organizes Workspace and Persona memories and prepares, updates, and injects subtask-level guidance throughout execution.
\item A trained memory-use policy and asynchronous collaboration between PersonaAgent and the main agent for preparing, updating, and injecting guidance.
\item Evaluation across three task settings, demonstrating gains in personalized task accuracy and execution efficiency.
\end{itemize}

\section{Related Work}
\paragraph{Long-term memory organization.}
Long-term memory systems retain information across interactions and make it available to later queries \citep{ref43,ref44,ref45}. MemoryBank stores conversation histories and summaries, updates user information over time, and incorporates a forgetting mechanism \citep{ref41}. Generative Agents combines observation, retrieval, and reflection to support later actions \citep{ref25}. MemGPT manages information across in-context and external memory \citep{ref24}, while MemOS provides a unified framework for organizing and managing agent memory \citep{ref17}. LangMem provides persistent-memory extraction, updating, and search tools \citep{ref15}; Mem0 supports scalable extraction, consolidation, and retrieval of interaction facts \citep{ref04}. A-MEM organizes memory into evolving, linked notes \citep{ref33}, and LightMem uses staged processing to reduce online maintenance costs \citep{ref05}. Dialogue-memory studies also examine historical context and reflective memory management across extended interactions \citep{ref10,ref30,ref47,ref19,ref32}. Together, these methods support historical information storage, organization, and retrieval \citep{ref46,ref48}.

\paragraph{Model collaboration.}
Multi-agent systems distribute complex tasks through role assignment, information exchange, and joint decision-making. AgentVerse studies dynamically organizing collaborative groups according to task requirements \citep{ref03}; AutoGen coordinates language models, tools, and human input through programmable multi-agent conversations \citep{ref49}; MetaGPT incorporates standard operating procedures into multi-role collaboration, organizing software-development tasks around structured intermediate outputs \citep{ref50}. These studies examine multi-agent collaboration through role organization, communication, and task workflows.

We apply model collaboration to personalized memory. PersonaAgent selects relevant memories and generates personalized guidance, while the main agent handles task planning and execution. Execution results inform memory and guidance updates between subtasks; asynchronous collaboration overlaps memory processing with task execution to support continuous personalization injection in long-horizon tasks.

\section{Method}
\label{sec:method}
\subsection{Overview}
\begin{figure*}[t]
\centering
\includegraphics[width=\textwidth]{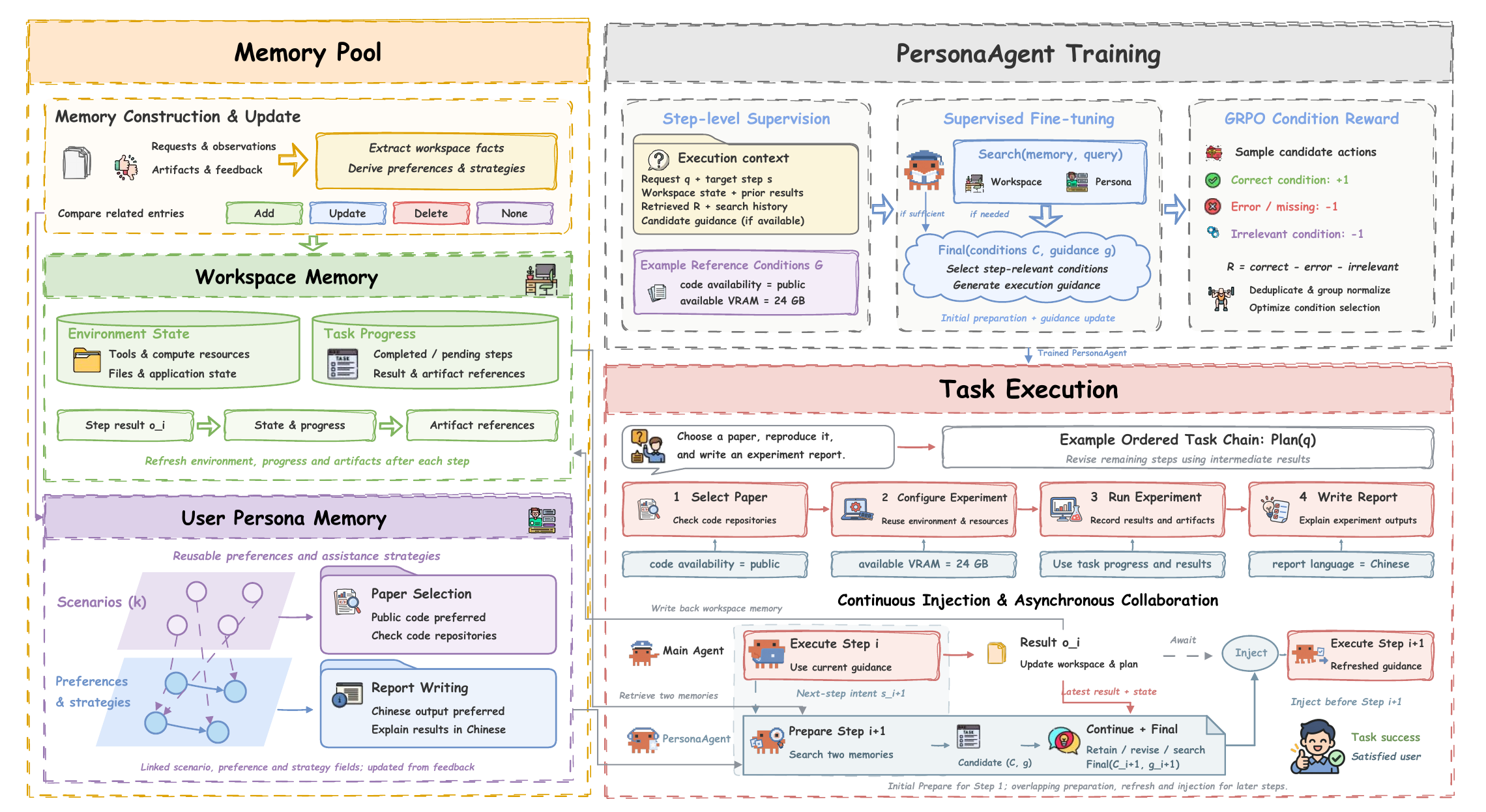}
\caption{AutoPersona's memory organization, policy training, and asynchronous execution. The main agent executes subtask $i$ while PersonaAgent retrieves both memory types and prepares guidance for subtask $i+1$. New results update the workspace and guidance before the next subtask begins.}
\label{fig:framework}
\end{figure*}

AutoPersona separates personalized memory use from task execution to enable continuous personalization injection. A main agent plans and advances an ordered task chain, while PersonaAgent retrieves relevant memory and converts it into personalized conditions and execution guidance. The framework combines two structured memory types, continuous subtask-level injection, and policy training. As shown in Figure~\ref{fig:framework}, Workspace Memory provides environment state, progress, and artifacts; Persona Memory provides preferences and assistance strategies. PersonaAgent prepares guidance for the next subtask during current execution and updates it when new results arrive.

\subsection{Two Types of Structured Memory}
We organize memory by information scope and update source:
\setcounter{equation}{0}
\begin{equation}
M=(M_W,M_P).
\label{eq:1}
\end{equation}

Each record in the historical interaction sequence $H=\{h_j\}_{j=1}^{N}$ is
\setcounter{equation}{1}
\begin{equation}
h_j=(q_j,\xi_j,A_j,f_j),
\label{eq:2}
\end{equation}

where $q_j$ is the user request, $\xi_j$ contains interactions and environment observations, $A_j$ denotes artifacts, and $f_j$ is user feedback or a task outcome. We extract task state and user preferences from these records to support subtask-level guidance.

\paragraph{Workspace Memory.}
Workspace Memory describes the current environment, constraints, and progress:
\setcounter{equation}{2}
\begin{equation}
M_W=(M_{\mathrm{env}},M_{\mathrm{task}}).
\label{eq:3}
\end{equation}

$M_{\mathrm{env}}$ stores tool configuration, computing resources, and file/application state. $M_{\mathrm{task}}$ stores task constraints, execution progress, previous-result summaries, and artifact references. For example, a Chinese-language requirement describes the current report; a selected paper and its public code describe a selection outcome; and environment preparation and result-file locations support subsequent execution. Workspace Memory is updated after each subtask. The prompt representation \code{workspace: \{env, task\}} directly corresponds to these two components (Appendices~\ref{app:memory} and~\ref{app:prompts}).

\paragraph{Persona Memory.}
Persona Memory stores reusable user preferences and assistance strategies. An item is
\setcounter{equation}{3}
\begin{equation}
p_l=(k_l,v_l,\sigma_l),
\label{eq:4}
\end{equation}

where $k_l$ is the applicable task or scenario, $v_l$ is the preference, and $\sigma_l$ is an assistance strategy. For example, a public-code preference suggests checking code availability during paper selection, whereas a preference for Chinese explanations guides report writing. Items are extracted from user expressions, choices, and feedback and updated through later interactions.

\paragraph{Construction and updating.}
The two memory extractors process each record as
\setcounter{equation}{4}
\begin{equation}
\begin{aligned}\Delta M_W^{(j)}&=F_W(h_j),\\\Delta M_P^{(j)}&=F_P\left(h_j,\Delta M_W^{(j)}\right).\end{aligned}
\label{eq:5}
\end{equation}

New information is compared with related items of the same type to choose an operation:
\setcounter{equation}{5}
\begin{equation}
\begin{aligned}\operatorname{op}&=F_{\mathrm{op}}(m,B,h_j),\\\operatorname{op}&\in\{\operatorname{Add},\operatorname{Update},\operatorname{Delete},\operatorname{None}\}.\end{aligned}
\label{eq:6}
\end{equation}

Here $m$ is newly extracted information and $B$ contains related retrieved items. The memory model adds new facts, updates changed states or preferences, deletes obsolete items, and skips duplicates. Execution outcomes mainly update Workspace Memory; user expressions and feedback update Persona Memory. Initial construction and online updating use the same extraction and storage-decision model.

\subsection{Continuous Injection Along a Task Chain}
\paragraph{Subtasks and personalized guidance.}
Given a user request $q$, the main agent plans
\setcounter{equation}{6}
\begin{equation}
\mathcal T=\langle s_1,s_2,\ldots,s_K\rangle
=\operatorname{Plan}(q),
\label{eq:7}
\end{equation}

Each $s_i$ is a subtask with a goal and operational requirements and may involve multiple tool operations. Subtasks execute in order, while intermediate results can revise the remaining plan. PersonaAgent supplies guidance before each subtask starts. Its execution context is
\setcounter{equation}{7}
\begin{equation}
x_i=(q,s_i,\omega_i,O_{<i}),
\qquad
O_{<i}=\langle o_1,\ldots,o_{i-1}\rangle,
\label{eq:8}
\end{equation}

where $\omega_i$ is the workspace state before execution and $O_{<i}$ contains preceding results, conveyed through summaries and artifact references. Retrieval and guidance generation are
\setcounter{equation}{8}
\begin{equation}
R_i=\operatorname{Retrieve}(M_i,x_i),
\label{eq:9}
\end{equation}

\setcounter{equation}{9}
\begin{equation}
(\mathcal C_i,g_i)=\operatorname{PersonaAgent}_\theta(x_i,R_i).
\label{eq:10}
\end{equation}

where $M_i$ is the current memory store. PersonaAgent uses \code{Search} to specify a memory type and query, and \code{Final} to return the condition set $\mathcal C_i$ and guidance $g_i$. Conditions record relevant keys and values; guidance organizes them for execution. The combined process is denoted $\operatorname{Prepare}_\theta(x_i,M_i)$. Action formats and retrieval settings appear in Appendix~\ref{app:memory}.

\paragraph{Injection and execution.}
Before subtask execution, we add conditions and guidance to its context:
\setcounter{equation}{10}
\begin{equation}
\begin{aligned}
s_i^+&=\operatorname{Inject}(s_i,\mathcal C_i,g_i),\\
o_i&=L_{\mathrm{main}}(q,s_i^+,\omega_i,O_{<i}).
\end{aligned}
\label{eq:11}
\end{equation}

Guidance changes with the subtask goal, while applicable preferences can be reused. The outcome updates Workspace Memory, yielding $\omega_{i+1}$ and $M_{i+1}$. The main agent aggregates completed results:
\setcounter{equation}{11}
\begin{equation}
r=\operatorname{Aggregate}(o_1,\ldots,o_K).
\label{eq:12}
\end{equation}

\paragraph{Asynchronous collaboration.}
When the next subtask intent is known, PersonaAgent uses
$\hat x_{i+1\mid i}=(q,s_{i+1},\omega_i,O_{<i})$ to prepare guidance concurrently with current execution:
\setcounter{equation}{12}
\begin{equation}
\begin{aligned}
o_i&\leftarrow L_{\mathrm{main}}(q,s_i^+,\omega_i,O_{<i}),\\
F_{i+1}&\leftarrow\operatorname{Async}\!\left[
\operatorname{Prepare}_\theta(\hat x_{i+1\mid i},M_i)
\right].
\end{aligned}
\label{eq:13}
\end{equation}

$F_{i+1}$ denotes the background preparation task. After subtask $i$, the system updates the workspace, and PersonaAgent combines the candidate with new results:
\setcounter{equation}{13}
\begin{equation}
\begin{aligned}(\hat{\mathcal C}_{i+1},\hat g_{i+1})&=\operatorname{Await}(F_{i+1}),\\(\mathcal C_{i+1},g_{i+1})&=\operatorname{Continue}_\theta\bigl(x_{i+1},M_{i+1};\\&\qquad\hat{\mathcal C}_{i+1},\hat g_{i+1}\bigr).\end{aligned}
\label{eq:14}
\end{equation}

$\operatorname{Continue}_\theta$ updates conditions and guidance and performs additional retrieval as needed. The first subtask uses direct preparation; plan revisions trigger preparation for the new intent. The full algorithm appears in Appendix~\ref{app:algorithms}.

Let $e_i$ be current execution time, $p_{i+1}$ the advance preparation time, $m_i$ memory-update and plan-revision time after execution, and $u_{i+1}$ guidance-update and injection time. When preparation and execution start together and remaining preparation overlaps the subsequent memory update, the inter-subtask wait is
\setcounter{equation}{14}
\begin{equation}
w_{i+1}=\max(m_i,p_{i+1}-e_i)+u_{i+1}.
\label{eq:15}
\end{equation}

The serial arrangement costs $m_i+p_{i+1}+u_{i+1}$ after execution. The asynchronous arrangement overlaps preparation with execution and memory updating.

\subsection{Training PersonaAgent}
Supervised fine-tuning supports instruction following and tool-use actions \citep{ref23,ref02}, while task-based optimization improves decisions in multi-turn tool interaction \citep{ref13}. PersonaAgent first learns \code{Search}/\code{Final} actions and guidance through SFT, then optimizes condition selection with GRPO. Scripts synthesize profiles, behavior histories, and web tasks following the Persona2Web construction process, producing 200 SFT and 1,000 RL examples. GPT-5.2-Codex annotates conditions and actions \citep{ref21}. A decision input is
\setcounter{equation}{15}
\begin{equation}
\begin{aligned}\chi&=(x,R,\eta,\hat D),\\x&=(q,s,\omega,O).\end{aligned}
\label{eq:16}
\end{equation}

where $x$ is the subtask context, $R$ contains retrieval results, $\eta$ records retrieval, and $\hat D=(\hat{\mathcal C},\hat g)$ contains candidate guidance. Initial preparation has no candidate; updating includes the candidate and new results. Each example has a reference condition set $\mathcal G=\{(k_j,v_j)\}$.

Teacher demonstrations retrieve memory with \code{Search} and output conditions and guidance with \code{Final}. SFT minimizes
\setcounter{equation}{16}
\begin{equation}
\mathcal L_{\mathrm{SFT}}(\theta)
=-\mathbb E_{(\chi,d^*)\sim\mathcal D_{\mathrm{SFT}}}
\log\pi_\theta(d^*\mid\chi).
\label{eq:17}
\end{equation}

where $d^*$ is the reference action and loss is computed over generated action tokens. Training covers guidance preparation and updating. Appendix~\ref{app:training} provides settings and data construction.

Starting from the SFT policy, we sample $G$ candidate action sequences per input and score their final condition sets:
\setcounter{equation}{17}
\begin{equation}
\begin{aligned}R(d)={}&N_{\mathrm{correct}}(d)\\&-N_{\mathrm{error}}(d)-N_{\mathrm{irrelevant}}(d).\end{aligned}
\label{eq:18}
\end{equation}

$N_{\mathrm{correct}}$, $N_{\mathrm{error}}$, and $N_{\mathrm{irrelevant}}$ count correct necessary conditions, erroneous or omitted conditions, and extra conditions unrelated to the target subtask. Predictions are normalized and matched to $\mathcal G$. GRPO uses group-normalized advantages, clipped policy ratios, and KL regularization against the SFT reference policy \citep{ref27}; the complete objective and counting rules are retained in Appendices~\ref{app:training} and~\ref{app:grpo}.

\section{Experiments}
\label{sec:experiments}
We evaluate preference and task-constraint satisfaction, task completion, and memory-use costs on Persona2Web, VitaBench~2.0, and \pibench{}. We compare six memory baselines and analyze components through progressive ablations.

\subsection{Experimental Setup}
\paragraph{Benchmarks.}
Persona2Web tests whether agents recover conditions such as website choice, size, and budget from user histories \citep{ref14}. Each query independently initializes memory from its designated history. VitaBench~2.0 evaluates preference application and updating in food delivery, local services, and online travel \citep{ref42}. We score the 668 subtasks whose \code{skill\_tested} excludes \code{proactive}, retain preference-update tasks, and execute each user's sequence chronologically. \pibench{} evaluates multi-session workflows with hidden needs and task dependencies \citep{ref37}. Methods start from the same initial environment and subsequently continue from their own preceding workspace and memory states.

\paragraph{Baselines and models.}
We compare Base+RAG \citep{ref16}, A-MEM \citep{ref33}, Mem0 (Entity) \citep{ref04}, LangMem \citep{ref15}, LightMem \citep{ref05}, and PersonaMem-v2 \citep{ref12}. All methods use GPT-5-mini as the main executor \citep{ref20}. AutoPersona also uses GPT-5-mini for memory construction and updating, and a trained Qwen3-8B PersonaAgent for retrieval and guidance \citep{ref35}. Methods share task order, permitted history, tool permissions, interaction budgets, and evaluation splits, with separate environment instances and memory stores. Training uses 200 script-synthesized SFT and 1,000 RL examples, GPT-5.2-Codex teacher annotations, LoRA SFT, and GRPO. VitaBench~2.0 and \pibench{} provide cross-dataset evaluation. Appendix~\ref{app:experiments} details protocols and adaptation.

\paragraph{Metrics and costs.}
Persona2Web reports website exact match (Web EM), preference recall (Pref R), and their mean (P-Avg). VitaBench~2.0 reports mean rubric satisfaction, the percentage of subtasks satisfying every trajectory criterion (Trajectory Success), and success of the entity or order produced by the last valid write (Outcome Success). Outcome Success uses valid outcome-scored subtasks; subtasks without writes are N/A. \pibench{} reports Checklist and Proactiveness scores for explicit deliverables and hidden needs, respectively, and their mean. All task metrics are percentages.

Memory storage output counts generated tokens from memory construction and write/update calls. Input counts memory or guidance tokens actually delivered to the executor, including repeated delivery. End-to-end time covers online retrieval, inference, tool calls, and retries, excluding offline indexing and judge scoring. Costs are per-task means; tokens are reported in thousands (K) and time in seconds. Shared initial-memory construction is amortized over a user's scored tasks.

\begin{table*}[t]
\centering
\footnotesize
\setlength{\tabcolsep}{1.8pt}
\renewcommand{\arraystretch}{1.18}
\caption{Main results on Persona2Web and VitaBench~2.0. Scores are percentages. Storage and Input are memory-maintenance output and memory/guidance input to the executor, in K tokens; Time is in seconds. Avg. is P-Avg; Traj. and Outcome denote trajectory and outcome success. Bold and underlining mark the best and second-best values. Base+RAG Storage is marked ``-'' and excluded from ranking. All methods use GPT-5-mini as executor; AutoPersona uses Qwen3-8B as PersonaAgent.}
\label{tab:main}
\begin{tabularx}{\textwidth}{@{}l*{12}{Z}@{}}
\toprule
\rowcolor{TableHeader}
\textbf{Method} & \multicolumn{6}{c}{\color{TableInk}\textbf{Persona2Web}} & \multicolumn{6}{c}{\color{TableInk}\textbf{VitaBench 2.0}}\\
\cmidrule(lr){2-7}\cmidrule(lr){8-13}
\rowcolor{TableHeader}
& \shortstack{Web\\EM$\uparrow$} & \shortstack{Pref\\R$\uparrow$} & Avg.$\uparrow$ & \shortstack{\scriptsize Storage\\$\downarrow$} & Input$\downarrow$ & Time$\downarrow$
& Rubric$\uparrow$ & Traj.$\uparrow$ & \shortstack{\scriptsize Outcome\\$\uparrow$} & \shortstack{\scriptsize Storage\\$\downarrow$} & Input$\downarrow$ & Time$\downarrow$\\
\midrule
\rowcolor{TableModel}
\multicolumn{13}{c}{\personamodel}\\
\midrule
% BEGIN DATA tables/main_rows
Base+RAG & 38.2 & 52.8 & 45.5 & - & 6.8 & \underline{57.6} & 57.8 & 21.3 & 27.1 & - & 9.6 & 236.4 \\
\rowcolor{TableStripe}A-MEM & 40.4 & 54.3 & 47.4 & 6.3 & 3.7 & 72.9 & 62.1 & 26.2 & 32.2 & 10.2 & 5.2 & 247.3 \\
Mem0 (Entity) & 59.6 & 69.5 & 64.6 & 3.9 & 2.9 & 69.3 & 65.3 & 29.2 & 35.2 & 7.6 & 4.4 & 231.6 \\
\rowcolor{TableStripe}LangMem & 56.4 & 60.4 & 58.4 & 4.2 & 3.4 & 67.7 & 64.2 & 28.3 & 34.1 & 8.3 & 4.8 & 238.2 \\
LightMem & 34.7 & 54.2 & 44.5 & \underline{1.7} & \underline{1.8} & 62.1 & 61.1 & 24.3 & 30.1 & \underline{3.7} & \underline{3.1} & \underline{222.7} \\
\rowcolor{TableStripe}PersonaMem-v2 & \underline{60.6} & \underline{76.9} & \underline{68.8} & 3.2 & 2.6 & 75.5 & \underline{67.4} & \underline{31.3} & \underline{37.3} & 6.2 & 4.1 & 253.8 \\
\rowcolor{TableHighlight}\textbf{AutoPersona} & \textbf{71.7} & \textbf{85.8} & \textbf{78.8} & \textbf{1.4} & \textbf{1.3} & \textbf{44.2} & \textbf{78.4} & \textbf{42.5} & \textbf{48.4} & \textbf{2.8} & \textbf{2.4} & \textbf{197.3} \\
% END DATA tables/main_rows
\bottomrule
\end{tabularx}
\end{table*}

\begin{table*}[t]
\centering
\small
\setlength{\tabcolsep}{4pt}
\renewcommand{\arraystretch}{1.18}
\caption{Main results on \pibench{}. Mean averages Checklist and Proactiveness. Scores are percentages; token and time units and highlighted values follow Table~\ref{tab:main}. AutoPersona uses Qwen3-8B as PersonaAgent.}
\label{tab:pi}
\begin{tabularx}{\textwidth}{@{}lrr*{4}{Z}@{}}
\toprule
\rowcolor{TableHeader}
\textbf{Method} & \textbf{Checklist}$\uparrow$ & \textbf{Proactiveness}$\uparrow$ & \textbf{Mean}$\uparrow$ & \textbf{Storage}$\downarrow$ & \textbf{Input}$\downarrow$ & \textbf{Time}$\downarrow$\\
\midrule
\rowcolor{TableModel}
\multicolumn{7}{c}{\personamodel}\\
\midrule
% BEGIN DATA tables/pibench_rows
Base+RAG & 54.3 & 65.9 & 60.1 & - & 8.7 & 239.5 \\
\rowcolor{TableStripe}A-MEM & 49.1 & 55.6 & 52.4 & 7.3 & 4.8 & 243.2 \\
Mem0 (Entity) & 50.5 & \underline{67.4} & 59.0 & 5.6 & 3.7 & 242.1 \\
\rowcolor{TableStripe}LangMem & 51.6 & 57.3 & 54.5 & 6.2 & 4.3 & 277.3 \\
LightMem & \underline{59.3} & 66.5 & \underline{62.9} & 4.8 & \underline{2.8} & \underline{229.4} \\
\rowcolor{TableStripe}PersonaMem-v2 & 56.4 & 65.1 & 60.8 & \underline{4.4} & 3.5 & 303.9 \\
\rowcolor{TableHighlight}\textbf{AutoPersona} & \textbf{67.2} & \textbf{71.3} & \textbf{69.3} & \textbf{3.1} & \textbf{2.1} & \textbf{199.7} \\
% END DATA tables/pibench_rows
\bottomrule
\end{tabularx}
\end{table*}

\subsection{Main Results}
\paragraph{Recovering historical preferences.}
On Persona2Web, AutoPersona achieves the best Web EM, Pref R, and P-Avg among all compared methods: 71.7, 85.8, and 78.8, respectively. These exceed the best baseline, PersonaMem-v2, by 11.1, 8.9, and 10.0 percentage points (Table~\ref{tab:main}). Improvements in both website selection and preference recall indicate gains in choosing key options and recovering user constraints.

\paragraph{Applying conditions to task execution.}
On VitaBench~2.0, AutoPersona achieves the best Rubric Satisfaction, Trajectory Success, and Outcome Success among all compared methods: 78.4, 42.5, and 48.4, respectively. These exceed the best baseline, PersonaMem-v2, by 11.0, 11.2, and 11.1 points. Gains in rubric satisfaction are accompanied by improved trajectory and outcome success. Complete-trajectory success remains lower than criterion-level satisfaction, showing the difficulty of satisfying several conditions simultaneously.

On \pibench{}, AutoPersona ranks first on all three metrics, with 67.2 Checklist, 71.3 Proactiveness, and 69.3 mean score (Table~\ref{tab:pi}). Checklist and mean score exceed the best baseline, LightMem, by 7.9 and 6.4 points; Proactiveness exceeds the best baseline, Mem0, by 3.9 points. AutoPersona improves satisfaction of both explicit deliverable criteria and hidden needs.

\paragraph{Token and time costs.}
Input is 1.3, 2.4, and 2.1 K on the three benchmarks, reducing input by 27.8\%, 22.6\%, and 25.0\% relative to LightMem. End-to-end times of 44.2, 197.3, and 199.7 seconds are 23.3\%, 11.4\%, and 12.9\% lower than the fastest baseline on each benchmark. Memory storage outputs of 1.4, 2.8, and 3.1 K are the lowest among methods included in this metric. Base+RAG Storage is marked ``-'' and excluded from its ranking; embedding and storage costs are accounted for separately. The main tables characterize the complete system, with component effects examined below.

The benchmarks complement each other: Persona2Web evaluates preference recovery, VitaBench~2.0 evaluates preference application in trajectories and outcomes, and \pibench{} evaluates cross-session deliverable standards and hidden needs.

\section{Ablation Studies}
\label{sec:ablations}
\begin{table*}[t]
\centering
\small
\caption{Progressive ablations grouped by memory, retrieval, and training. P2W and \pibench{} averages follow Tables~\ref{tab:main} and~\ref{tab:pi}; Vita averages its three task metrics. Overall averages the three benchmarks. Costs are benchmark-macro-averaged per-task means. Structured Memory and Query-only Injection are the same reference configuration and share results.}
\label{tab:ablation}
\begin{tabular}{llrrrrrrr}
\toprule
\rowcolor{TableHeader}
Group & Configuration & P2W$\uparrow$ & Vita$\uparrow$ & $\pi$-Bench$\uparrow$ & Overall$\uparrow$ & Storage$\downarrow$ & Input$\downarrow$ & Time$\downarrow$\\
\midrule
% BEGIN DATA tables/ablation_rows
 & Only W & 57.1 & 34.2 & 52.8 & 48.0 & 1.5 & 0.9 & 129.6 \\
\rowcolor{TableStripe} & Only P & 61.8 & 38.4 & 55.3 & 51.8 & 1.6 & 1.0 & 131.8 \\
 & Flat Memory & 63.4 & 42.1 & 58.1 & 54.5 & 3.9 & 2.2 & 145.7 \\
\rowcolor{TableStripe}\multirow{-4}{*}{Memory} & Structured Memory & 65.2 & 44.0 & 58.6 & 55.9 & 2.0 & 1.2 & 132.4 \\
 & Query-only Injection & 65.2 & 44.0 & 58.6 & 55.9 & 2.0 & 1.2 & 132.4 \\
\rowcolor{TableStripe} & Parallel K Injection & 66.7 & 45.2 & 59.1 & 57.0 & 2.2 & 2.7 & 153.6 \\
\multirow{-3}{*}{Retrieval} & Sequential Injection & 68.2 & 46.3 & 59.8 & 58.1 & 2.3 & 2.5 & 185.4 \\
\rowcolor{TableStripe} & SFT & 69.8 & 47.9 & 60.9 & 59.5 & 2.3 & 2.5 & 179.2 \\
\rowcolor{TableHighlight}\multirow{-2}{*}{Training} & \textbf{SFT + RL (Full)} & \textbf{78.8} & \textbf{56.4} & \textbf{69.3} & \textbf{68.2} & \textbf{2.4} & \textbf{1.9} & \textbf{147.1} \\
% END DATA tables/ablation_rows
\bottomrule
\end{tabular}
\end{table*}

\paragraph{Design.}
Table~\ref{tab:ablation} progressively compares memory sources and organization, retrieval and injection, and policy training. Configurations share evaluation splits, initial histories and environments, the executor, planning, tool permissions, and task budgets. Structured Memory and Query-only Injection reuse one reference configuration. Full combines structured memory, adaptive \code{Search}/\code{Final}, continuous injection, asynchronous collaboration, and SFT+GRPO. Appendix~\ref{app:ablation} specifies components and comparison targets.

\paragraph{Memory sources and organization.}
Only W and Only P obtain Overall scores of 48.0 and 51.8, versus 55.9 for Structured Memory. Combining task state with cross-task preferences therefore improves over either source alone. With the same source information, Structured Memory exceeds Flat Memory by 1.4 points and reduces storage output from 3.9 to 2.0 K, input from 2.2 to 1.2 K, and time from 145.7 to 132.4 seconds.

\begin{figure}[t]
\centering
\includegraphics[width=\columnwidth]{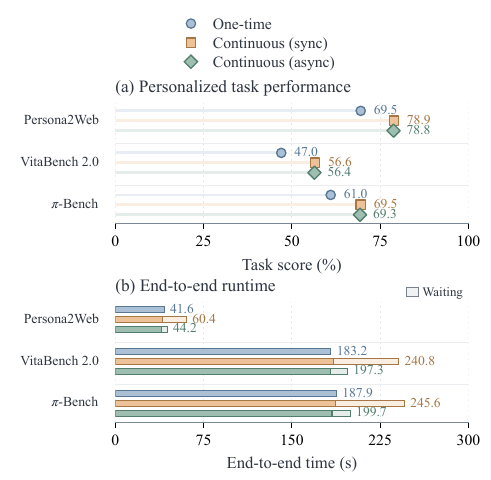}
\caption{Injection and scheduling with a fixed trained PersonaAgent. (a) Task scores for the three configurations. (b) End-to-end time: execution (solid) and subtask-boundary waiting (pale).}
\label{fig:asynchronous}
\end{figure}

\paragraph{Injection and training.}
Parallel K and Sequential Injection improve over Query-only Injection by 1.1 and 2.2 Overall points. Sequential Injection achieves 58.1, but increases input to 2.5 K and time to 185.4 seconds. Adding SFT raises Overall to 59.5; Full reaches 68.2, gaining 8.7 points over SFT. Full reduces input to 1.9 K and time to 147.1 seconds. Relative to Sequential Injection, Full improves Overall by 10.1 points and reduces time by 20.7\%. Figure~\ref{fig:asynchronous} separately examines injection frequency and scheduling with a fixed trained policy.

\begin{figure}[t]
\centering
\includegraphics[width=\columnwidth]{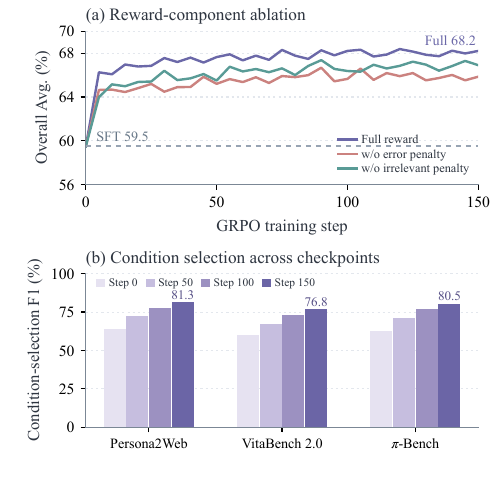}
\caption{Reward and condition-selection analysis. (a) Overall task scores during GRPO under three reward designs. (b) Condition-selection F1 at four checkpoints. SFT and Full scores are 59.5 and 68.2 (Table~\ref{tab:ablation}).}
\label{fig:training}
\end{figure}

\paragraph{Reward design and condition selection.}
Figure~\ref{fig:training}(a) compares the full reward with removing $N_{\mathrm{error}}$ or $N_{\mathrm{irrelevant}}$. Removing the error term removes penalties for omissions, incorrect values, and inconsistent conditions; removing the irrelevant term affects extra conditions that are correct but unrelated to the subtask. Figure~\ref{fig:training}(b) tracks condition-selection F1 after normalizing and matching condition keys and values to the reference set. Precision is the fraction of selected conditions that are correct and necessary; recall is the fraction of reference conditions correctly covered, with $F1=2PR/(P+R)$. This metric evaluates condition selection, while task scores evaluate the executor's final results.

\paragraph{Continuous injection and asynchronous scheduling.}
We fix the trained PersonaAgent, memory system, and main executor and compare three configurations. One-time injection supplies guidance at task initialization. Continuous synchronous injection prepares and updates guidance before each subtask, with execution waiting for this process. Continuous asynchronous injection prepares the next subtask's guidance during current execution and updates it when new results arrive. Figure~\ref{fig:asynchronous} reports task scores and end-to-end times using the benchmark metrics defined above; Appendix~\ref{app:scheduling} details the configurations.

Continuous synchronous injection raises scores from 69.5 to 78.9 on Persona2Web, 47.0 to 56.6 on VitaBench~2.0, and 61.0 to 69.5 on \pibench{} relative to one-time injection, gaining 9.4, 9.6, and 8.5 points. Asynchronous injection scores 78.8, 56.4, and 69.3, respectively, which are 0.1, 0.2, and 0.2 points below synchronous injection. Its times are 44.2, 197.3, and 199.7 seconds, compared with 60.4, 240.8, and 245.6 seconds for synchronous injection: reductions of 26.8\%, 18.1\%, and 18.7\%. Subtask-boundary waits fall from 20.6, 55.4, and 58.7 seconds to 5.3, 14.7, and 15.6 seconds. One-time injection takes 41.6, 183.2, and 187.9 seconds, running faster but yielding lower task scores. Continuous injection improves task scores, while asynchronous scheduling reduces its time cost with score differences of at most 0.2 points from synchronous injection.

Boundary waiting falls by 74.3\%, 73.5\%, and 73.4\%, accounting for over 93\% of total time savings. Non-waiting time changes modestly, supporting guidance--execution overlap as the main source of speedup.

\section{Conclusion}
AutoPersona continuously injects personalization into long-horizon tasks using Workspace Memory for task state and Persona Memory for reusable preferences. A PersonaAgent trained with SFT and GRPO retrieves relevant information and generates guidance throughout an ordered task chain. Asynchronous preparation and result-dependent updating align personalization with evolving task state. Evaluation on Persona2Web, VitaBench~2.0, and \pibench{} shows gains in task performance. Ablations support structured memory and continuous injection; with a fixed trained policy, asynchronous scheduling reduces end-to-end time by 18.1--26.8\% relative to synchronous injection.

\label{main-text-end}
\clearpage
\section*{Limitations}
Although AutoPersona demonstrates effectiveness across three benchmarks, its generalizability to broader task domains, user populations, and preference types remains to be further explored. The PersonaAgent is trained on synthetic data following the Persona2Web construction process, and its performance depends on the quality of memory extraction and retrieval. In more complex real-world scenarios, noisy interaction histories, ambiguous preferences, or evolving user needs may affect the accuracy of retrieved information and the resulting personalized guidance. Further evaluation on diverse real-world interaction histories is therefore needed to establish the robustness and generalizability of the learned memory-use policy. In addition, the asynchronous collaboration mechanism relies on a sufficiently clear next-subtask intent to prepare personalized guidance in advance. When intermediate execution results lead to substantial changes in the remaining task plan, previously prepared guidance may need to be revised or discarded, introducing additional overhead. Consequently, workflows involving frequent plan revisions or very short subtasks may benefit less from asynchronous preparation, as the costs of guidance preparation and updating can partially offset the latency savings.

\section*{Ethical Considerations}
Personalized memory can contain sensitive user information. Memory access is organized by user and workspace scope, and persistent storage should support appropriate retention and deletion. The evaluation follows the interaction protocols of the supplied benchmarks.

\bibliography{references}
\clearpage
\appendix
\section{Experimental Details}
\label{app:experiments}
\subsection{Training and Evaluation Settings}
\label{app:training}
\paragraph{Training data.}
Following the Persona2Web construction process \citep{ref14}, scripts synthesize profiles, events, web actions, chronologically organized behavior histories, and task queries with three levels of preference explicitness. The resulting training data contain 200 SFT and 1,000 RL examples. Each example includes a history, current request, target subtask, and extracted memories. SFT supervises \code{Search}/\code{Final}, condition selection, and guidance. RL scores final condition sets against reference conditions. Persona2Web evaluates personalized web tasks, while VitaBench~2.0 and \pibench{} provide cross-dataset evaluation.

\paragraph{Model responsibilities.}
GPT-5-mini extracts Workspace and Persona memories and selects Add, Update, Delete, or None. A Qwen3-8B policy trained with SFT and GRPO serves as PersonaAgent. GPT-5-mini plans and executes tasks on all three benchmarks. GPT-5.2-Codex generates reference conditions, retrieval actions, guidance, and guidance-update demonstrations; it also serves as the VitaBench~2.0 judge. Persona2Web and \pibench{} use benchmark labels, rules, or official evaluation procedures.

The teacher receives the request, target subtask, workspace state, preceding outputs, and relevant memory, and annotates $\mathcal G=\{(k_j,v_j)\}$. It merges equivalent conditions. Initial-preparation examples contain no candidate guidance; update examples add the candidate and newly completed results. The teacher demonstrates extra retrieval when required and returns conditions and guidance through \code{Final}. During RL, the current policy generates rollouts using vLLM. Reference labels supply supervision and scoring, while policy inputs contain request, subtask state, preceding results, and retrieval returns. Appendix~\ref{app:prompts} provides the prompts.

\begin{table}[t]
\centering\small
\caption{LoRA SFT configuration.}
\label{tab:sft}
\begin{tabularx}{\columnwidth}{YX}
\toprule
Setting & Value\\\midrule
Backbone & Qwen/Qwen3-8B\\
Framework & LLaMA-Factory\\
Fine-tuning & LoRA\\
Rank / alpha & 8 / 16\\
Per-device batch & 1\\
Gradient accumulation & 8\\
Learning rate & $1\times10^{-6}$\\
Training data & Script-synthesized, following Persona2Web\\
Examples & 200\\
Actions & Search / Final\\
Epochs & 2\\
Precision & BF16\\
Checkpoint interval & 50 steps\\
\bottomrule
\end{tabularx}
\end{table}

\begin{table}[t]
\centering\small
\caption{GRPO configuration.}
\label{tab:rl}
\begin{tabularx}{\columnwidth}{YX}
\toprule
Setting & Value\\\midrule
Framework / algorithm & VERL 0.7.0 / GRPO\\
Hardware & $8\times$ NVIDIA H100-80GB\\
Rollout engine & vLLM\\
Group size & 8\\
Rollouts per query & 8\\
Learning rate & $1\times10^{-6}$\\
Training data & Script-synthesized, following Persona2Web\\
Examples & 1,000\\
Training steps & 150\\
Evaluation / checkpoint & Every 5 / 20 steps\\
Seeds & 42, 43, 44\\
\bottomrule
\end{tabularx}
\end{table}

\paragraph{Condition scoring.}
Fields, units, and synonymous expressions are normalized before prediction/reference matching, and duplicate outputs are merged. $N_{\mathrm{correct}}$ counts necessary conditions with correct values. $N_{\mathrm{error}}$ counts reference conditions that are not correctly covered and extra conditions with incorrect values, conflicting values for one key, or inconsistency with available information. An incorrect value and omission for the same reference item count once. $N_{\mathrm{irrelevant}}$ counts extra conditions whose content is correct but does not affect the target subtask. Relevance follows the subtask goal and annotations. Guidance expression is learned through SFT. The reward remains Equation~\ref{eq:18}.

\paragraph{Interaction protocols.}
The executor handles business dialogue and environment tools; PersonaAgent's internal Search retrieves memory and Final supplies conditions and guidance. These processes are logged separately. In Persona2Web, each query uses its designated historical context independently and adds no new user questions. Multi-turn behavior occurs in retrieval and web execution. Inferences made during execution stay within the current task; later independent queries reinitialize from their own histories. VitaBench~2.0's official PersonalizationUser supplies business replies, such as order confirmations, with common simulator settings and interaction budgets across methods. The executor and memory module receive visible requests, responses, and observations. In \pibench{}, the official user simulator advances sessions using executor replies and workspace artifacts; PersonaAgent uses messages and outcomes already available in history.

\paragraph{Scoring and cost accounting.}
VitaBench~2.0 separately computes \code{traj\_reward}, \code{outcome\_reward}, and rubric satisfaction with GPT-5.2-Codex. Method names are hidden and candidate outputs randomized for judging. Persona2Web and \pibench{} use their benchmark evaluation procedures.

Storage output includes tokens returned by memory construction and write/update calls, including structured formatting, merge decisions, and reasoning tokens already counted by the service; reasoning tokens are not counted twice. Final stored-text size is separate. Input includes memory/guidance fragments actually delivered to the executor; repeated delivery is counted each time. Generic prompts, dialogue, and tool observations are excluded from that metric. Retrieval decisions, guidance generation, executor output, planning, user simulation, and judging are tracked separately, and fragment token counts should record the tokenizer and any estimation. End-to-end time includes online retrieval, PersonaAgent/executor inference, tool calls, and retries, excluding offline indexing and judge scoring.

An initial user-history memory is constructed once for a chronological sequence, and generation output is amortized over that user's scored subtasks. Incremental-history processing and online maintenance are assigned to their respective subtasks. Costs are averaged per task within each benchmark; Table~\ref{tab:ablation} then macro-averages benchmarks. Unscored-task and warm-up maintenance is separate. An independent single-query run includes complete initial construction. Retry usage is accumulated; missing usage remains unknown or partial. Base+RAG directly indexes raw history without generative memory maintenance. Its Storage is marked ``-'' and excluded from that metric; embedding, storage, and retrieval costs are accounted for separately.

\subsection{Datasets and Memory Context}
\label{app:datasets}
\paragraph{Persona2Web.}
The benchmark contains 50 user profiles, each with approximately one year of browsing history. Its 150 base tasks are instantiated at three levels of preference explicitness, yielding 450 queries \citep{ref14}. Memory is extracted from browsing and purchasing records preceding each test task. Web EM measures exact website match. Pref R averages query-level recall of annotated preference values; P-Avg averages Web EM and Pref R.

\paragraph{VitaBench 2.0.}
The full benchmark contains 56 users and 771 subtasks involving food delivery, local services, and online travel \citep{ref42}. A user's chronological sequence includes requests, past interactions, candidate environments, and scoring criteria. The \code{update} label tests addition, deletion, or modification of preferences and their later use; \code{proactive} tests asking users or exploring the environment to obtain information. Labels can co-occur. The fixed data version contains 147 update subtasks, including 16 also labeled proactive. We maintain memory in the original task order and score the 668 non-proactive subtasks: 131 update and 537 non-update subtasks.

Rubric Satisfaction averages the fraction of criteria met per scored subtask. Trajectory Success counts subtasks satisfying every criterion. Outcome scoring is enabled with \code{--enable-outcome-reward}. Outcome Success evaluates the entity or order from the last valid write and includes only valid outcome-scored subtasks; no-write subtasks are N/A. Per-method valid sample counts are required for reporting this metric. Vita averages the three percentage-valued metrics.

\paragraph{\pibench{}.}
The benchmark contains 100 multi-turn tasks across five professional personas, with 20 tasks per persona \citep{ref37}. Checklist measures explicit task standards and Proactiveness measures hidden requirements; mean score averages the two. Tasks follow official user order. Files, application state, and artifacts persist across sessions, while each method maintains its own run-specific memory. A new user or independent repeat starts from the initial environment.

\subsection{Baseline Adaptation}
\label{app:baselines}
Base+RAG indexes visible history; A-MEM builds dynamic links; Mem0 (Entity) organizes entities and relations; LangMem integrates long-term memories; LightMem uses staged compression; PersonaMem-v2 applies its user-memory policy. All are connected to the same GPT-5-mini executor and official \pibench{} tools through common read/write interfaces.

Write records include visible requests/replies, completed actions, tool returns, artifact references, and user/session/time/source identifiers. The adapter converts these records to each baseline's native input while preserving its memory organization and updating algorithm. Reads query with the current request, subtask intent, and visible state and provide returned items or summaries to the executor. PersonaMem-v2's policy maintains user memory; the benchmark user simulator supplies replies.

After each subtask, observed interactions and outcomes are written through the adapter. Before the next subtask, baselines retrieve according to the request and intent; AutoPersona prepares and updates guidance as in Appendix~\ref{app:algorithms}. Memory persists between sessions, with files and applications carried forward by the official environment. Methods and repeats have separate environments and memory stores, sharing initial state, history, task order, executor, planning, tools, and budgets. Returned memory uses a common context-length budget and retains native ordering/organization. Queries, writes, model/tool calls, and user replies are logged separately. Vector-based memory retrieval uses OpenAI's \code{text-embedding-3-small}. PersonaAgent uses a 16K-token input-context limit for both guidance preparation and updating.

\subsection{Progressive Ablation Settings}
\label{app:ablation}
The memory group compares Only W, Only P, Flat Memory, and Structured Memory. Flat and Structured use the same source content; the former indexes it together and the latter organizes it by memory type. Structured Memory retrieves at query initialization and injects once. Query-only Injection is the identical configuration reused as the retrieval-group reference.

Parallel K Injection adds parallel retrieval/injection; Sequential Injection uses sequential retrieval and subtask-level injection. SFT trains PersonaAgent on the preceding memory/execution flow. Full combines structured memory, adaptive Search/Final, continuous injection, asynchronous collaboration, and SFT+GRPO. Configurations share splits, initial history/environment, executor, planning, tool permissions, and budgets, and update their own environment/memory using their observed outcomes. Stage comparisons measure combined configuration effects.

Additional retrieval is capped at $k=3$, counted after common initial retrieval. Each Search in this comparison contains one memory/query pair. Parallel queries are generated independently from the same initial context and their results merged. Adaptive retrieval uses new results to decide whether to search again or return Final. Ordering, deduplication, and memory budgets are shared.

\subsection{Injection and Scheduling Comparison}
\label{app:scheduling}
Figure~\ref{fig:asynchronous} fixes the trained Qwen3-8B PersonaAgent, the Workspace/Persona memory system, and the GPT-5-mini executor. One-time injection provides guidance at task initialization and adds no new guidance at later subtask boundaries. Continuous synchronous injection performs preparation and result-dependent updating at each boundary before execution proceeds. Continuous asynchronous injection overlaps preparation for the next subtask with current execution and incorporates the completed results before injection. The synchronous and asynchronous configurations use the same guidance policy and differ in scheduling.

Task scores use P-Avg on Persona2Web, the mean of the three VitaBench~2.0 task metrics, and the mean of Checklist and Proactiveness on \pibench{}. Time is mean online end-to-end wall-clock time per task. Solid bar segments (Execution) show the non-waiting portion; pale segments (Waiting) show waiting at subtask boundaries. Their sum is the total end-to-end time (Total). Preparation that overlaps execution is represented within wall-clock time, rather than added again as an independent duration. Task scores and times in the figure are shown to one decimal place.

\subsection{Personalized Food-Delivery Case}
\label{app:case}
\begin{figure*}[t]
\centering
\includegraphics[width=\textwidth]{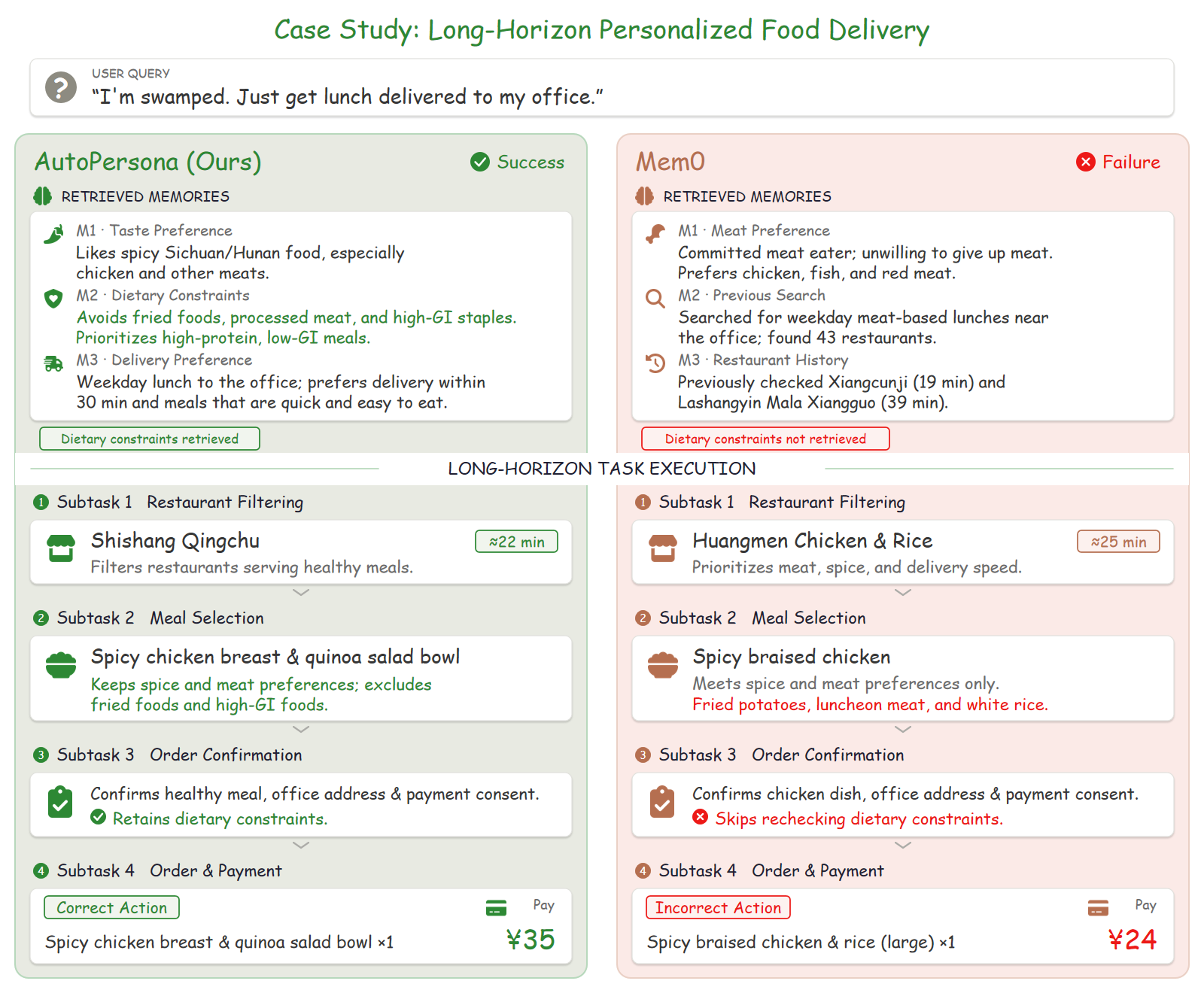}
\caption{Personalized food delivery from a cloud execution case. AutoPersona and Mem0 apply historical information during restaurant filtering, meal selection, order confirmation, and payment. The current request asks only for lunch delivery to the office.}
\label{fig:case}
\end{figure*}
AutoPersona recovers preferences for spicy food and meat, avoidance of fried and processed foods, and weekday delivery to the office within 30 minutes. These conditions guide subsequent filtering and actions. A spicy chicken-breast and quinoa salad bowl satisfies taste and dietary requirements; order confirmation also addresses the office location and payment consent. Mem0 recovers meat preferences and earlier restaurant searches but misses dietary conditions. Its order includes fried potatoes, luncheon meat, and white rice. This case illustrates how recovering only part of a profile can lead to an action that violates user requirements, and how subtask guidance carries relevant historical conditions through execution.

\subsection{Human Agreement with the Judge}
We sample 100 non-update and 100 update tasks from VitaBench~2.0. Human annotators independently apply the same trajectory and outcome criteria as GPT-5.2-Codex, with hidden method names and randomized candidate order. There are 95 and 96 agreements in the two groups (Table~\ref{tab:judge}). Agreement is the fraction of matching judgments. The standard deviation is computed from per-example binary agreement, using denominator $n-1$, on the 0/1 scale.

\begin{table}[t]
\centering\small
\caption{GPT-5.2-Codex and human agreement on VitaBench~2.0.}
\label{tab:judge}
\begin{tabularx}{\columnwidth}{Yrrr}
\toprule
Task type & $n$ & Agreement (\%) & SD\\\midrule
No update/proactive & 100 & 95.0 & 0.219\\
Update, no proactive & 100 & 96.0 & 0.197\\
Overall & 200 & 95.5 & 0.208\\
\bottomrule
\end{tabularx}
\end{table}

\subsection{Long-Horizon Personalization Consistency}
\label{app:long_horizon}

\begin{table*}[t]
\centering\small
\setlength{\tabcolsep}{6pt}
\renewcommand{\arraystretch}{1.18}
\caption{Long-horizon personalization consistency on VitaBench~2.0. Scores are percentages; higher is better. Early, Middle, and Late denote stages of each user's chronological subtask sequence. Overall reports the score across all stages.}
\label{tab:long_horizon}
\begin{tabularx}{\textwidth}{ll*{4}{Z}}
\toprule
\rowcolor{TableHeader}
\textbf{Metric} & \textbf{Method} & \textbf{Early} & \textbf{Middle} & \textbf{Late} & \textbf{Overall}\\
\midrule
\cellcolor{white} & Mem0 (Entity) & 68.2 & 65.4 & 62.3 & 65.3\\
\rowcolor{TableStripe}\cellcolor{white} & One-time Injection & 73.8 & 71.5 & 69.2 & 71.5\\
\rowcolor{TableHighlight}\cellcolor{white}\multirow{-3}{*}{Rubric Satisfaction$\uparrow$} & \textbf{AutoPersona} & \textbf{79.2} & \textbf{78.5} & \textbf{77.5} & \textbf{78.4}\\
\midrule
\cellcolor{white} & Mem0 (Entity) & 31.7 & 29.2 & 26.8 & 29.2\\
\rowcolor{TableStripe}\cellcolor{white} & One-time Injection & 36.6 & 34.9 & 33.2 & 34.9\\
\rowcolor{TableHighlight}\cellcolor{white}\multirow{-3}{*}{Trajectory Success$\uparrow$} & \textbf{AutoPersona} & \textbf{43.6} & \textbf{42.7} & \textbf{41.2} & \textbf{42.5}\\
\bottomrule
\end{tabularx}
\end{table*}

To further examine AutoPersona's ability to maintain personalization over extended user interactions, we conduct a long-horizon personalization consistency experiment on VitaBench~2.0. For each user, we partition the chronological subtask sequence into early, middle, and late stages, maintaining and updating memory throughout execution. We compare AutoPersona with Mem0 (Entity) and One-time Injection, which supplies personalization guidance only at task initialization. We report Rubric Satisfaction and Trajectory Success separately for each stage, following the execution and evaluation protocol in Appendix~\ref{app:datasets}.

Table~\ref{tab:long_horizon} shows that Mem0 and One-time Injection decline as interactions progress, while AutoPersona maintains comparatively stable \mbox{performance} across the three stages. This trend suggests a potential advantage of continuous injection: dynamically retrieving relevant preferences at successive subtask stages and generating personalized guidance using the evolving task state may reduce the risk that important preferences are not effectively used during long-term interactions.

\section{Memory and Execution Details}
\label{app:memory}
\subsection{Information Scope and Extraction}
Workspace Memory stores the current task environment, constraints, progress, and artifacts. Persona Memory stores reusable preferences and assistance strategies. For example, a preference for Chinese research reports belongs to the persona, while the current report's language, results, and path belong to the workspace. A public-code preference is reusable; the selected paper and code-check outcome are task-specific. Historical records can support both a reusable preference and its current-task outcome.

Initial memory construction starts from empty Workspace and Persona stores and processes historical records chronologically. $F_W$ and $F_P$ are implemented by GPT-5-mini in a joint structured response. It first organizes workspace information and then summarizes preferences using that information and the source record. Online results use the same extraction and updating process.

Workspace content uses \code{env} and \code{task}, corresponding to $M_{\mathrm{env}}$ and $M_{\mathrm{task}}$. Environment fields cover tool configuration, computing resources, and file/application state. Task fields cover the goal, constraints, completed/pending subtasks, previous-result summaries, and artifacts. Interaction and operation summaries are placed in the corresponding environment fields or \code{previous\_results}. Persona fields \code{topic}, \code{preference}, and \code{strategy} correspond to $k_l$, $v_l$, and $\sigma_l$.

\begin{lstlisting}
{
  "workspace": {
    "env": {
      "tool_config": {"python_environment": "ready"},
      "compute_resources": {"available_gpu_memory_gb": 24},
      "file_state": {"notes.md": "available"},
      "application_state": {}
    },
    "task": {
      "goal": "Reproduce a paper and write a report",
      "constraints": {"report_language": "Chinese"},
      "completed_subtasks": ["prepare"],
      "pending_subtasks": ["run_experiments", "write_report"],
      "previous_results": ["Environment prepared", "Report language confirmed"],
      "artifacts": {"notes": "notes.md"}
    }
  },
  "persona": [{
    "topic": "Research reports",
    "preference": "Explain findings in Chinese",
    "strategy": "Use Chinese and explain the main experimental findings"
  }]
}
\end{lstlisting}
Each stored item also has an ID, user/workspace scope, source-record ID, and update time. The persona is a list of extracted preferences. The memory model compares each new item with up to three positive-similarity candidates of the same type and scope. An empty candidate set yields Add. Otherwise it selects Add, Update, Delete, or None and returns the target and proposed content. Update combines new information with still-valid content; Delete removes obsolete items; None leaves duplicates unchanged. Workspace items are scoped to the current workspace, while a user's Persona items can be reused across that user's workspaces.

\subsection{Retrieval and Actions}
PersonaAgent searches \code{workspace} and \code{persona}. Initial retrieval queries both with the user request plus subtask instruction. Subsequent Search actions add targeted queries. Queries and memory items are encoded with OpenAI's \code{text-embedding-3-small}. Cosine similarity ranks candidates, with newer items first on ties.

\begin{table}[t]
\centering\small
\caption{Memory retrieval settings.}
\label{tab:retrieval}
\begin{tabularx}{\columnwidth}{YX}
\toprule
Setting & Value\\\midrule
Memory types & Workspace / Persona\\
Embedding model & {\scriptsize\code{text-embedding-3-small}}\\
Queries per Search & One or more memory/query pairs\\
Ranking & Cosine similarity; recency breaks ties\\
Returns & At most top-3 per memory type after ID deduplication\\
PersonaAgent input context & 16K tokens\\
Additional retrieval & At most 3 Search rounds per preparation/update\\
\bottomrule
\end{tabularx}
\end{table}
Non-positive matches are not returned. Memory items are included in rank order within PersonaAgent's input-context budget, alongside prompts, the request, and preceding results. Retrieved items include IDs, content, sources, and update times. Search actions return queries; Final returns \code{memory\_points} and \code{answer}, corresponding to $\mathcal C_i$ and $g_i$:

\begin{lstlisting}
{"action":"search", "searches":[
  {"memory":"persona", "query":"Research-report language preference"},
  {"memory":"workspace", "query":"Current experiment results and report paths"}
]}

{"action":"final", "memory_points":[
  {"key":"report_language", "value":"Chinese", "evidence_ids":["persona-001"]}
], "answer":"Explain the findings in Chinese and retain key English terminology."}
\end{lstlisting}
Sources use memory IDs, \code{current\_request}, \code{workspace\_state}, or \code{step\_result:<subtask-name>}. A preparation begins with initial retrieval and then invokes PersonaAgent. Each Search merges queries and re-applies top-3 and the memory budget; Final ends preparation. Three additional Search rounds allow at most four policy decisions. Continue uses the same loop and budget. The executor handles business tools and benchmark user interaction.

\subsection{Execution Algorithms}
\label{app:algorithms}
The main agent plans and executes ordered subtasks. Each subtask may involve multiple tool operations. PersonaAgent prepares the next subtask against a read-only memory snapshot during current execution. Completed outcomes update memory and state, and the same policy combines the candidate with new results before injection. Plans revise the unexecuted part of the chain. Execution state and committed results support resumption; tool effects are tracked through the tool layer. The serial arrangement uses the same preparation, memory update, and Continue but schedules candidate preparation after current execution.

Algorithms~\ref{alg:memory-update}--\ref{alg:prepare} describe memory updating, asynchronous execution, and the shared preparation loop. In Algorithms~\ref{alg:async-execution} and~\ref{alg:prepare}, $r_i=(q,V[i],V,\omega,O)$ packages the current request context, and $D_i=(\mathcal C_i,g_i)$ contains conditions and guidance. Retrieval calls use the supplied settings and scope; AsyncPrepare starts the Prepare phase.

\begin{algorithm}[tbp]
\renewcommand{\thealgorithm}{B1}
\caption{Memory extraction and updating}
\label{alg:memory-update}
\small
\begin{algorithmic}[1]
\Require Record $h$; memories $M_W,M_P$; memory model; retrieval settings and user/workspace scope
\Ensure Updated memories and operation log $L$
\State $(w,P) \gets \Call{Extract}{h}$
\State Store the source record for $h$; $L \gets [\,]$
\ForAll{$(k,m) \in \{(W,w)\}\cup\{(P,p):p\in P\}$}
  \State $C \gets \Call{RetrieveSameType}{m,M_k}$
  \If{$C=\varnothing$}
    \State $(a,t,m') \gets (\mathrm{Add},\varnothing,m)$
  \Else
    \State $(a,t,m') \gets \Call{Resolve}{m,C}$
  \EndIf
  \If{$a=\mathrm{Add}$}
    \State \Call{AddMemory}{$M_k,m',h$}
  \ElsIf{$a=\mathrm{Update}$}
    \State \Call{UpdateMemory}{$M_k,t,m',h$}
  \ElsIf{$a=\mathrm{Delete}$}
    \State \Call{DeleteMemory}{$M_k,t$}
  \Else
    \State Retain existing memory
  \EndIf
  \State Append $(k,a,t)$ to $L$
\EndFor
\State \Return $(M_W,M_P,L)$
\end{algorithmic}
\end{algorithm}

\begin{algorithm}[tbp]
\renewcommand{\thealgorithm}{B2}
\caption{Asynchronous personalized execution}
\label{alg:async-execution}
\small
\begin{algorithmic}[1]
\Require Request $q$; state $\omega$; memory $M$; main agent, PersonaAgent, memory model, and tools
\Ensure Aggregated subtask outputs
\State $V \gets \Call{OrderedPlan}{q}$; $O \gets$ committed results
\State $i \gets$ index of first uncommitted subtask
\State $r_i \gets (q,V[i],V,\omega,O)$
\State $D_i \gets \Call{Prepare}{r_i,M,\mathrm{Prepare}}$
\While{uncommitted subtasks remain}
  \State $v_i^+ \gets \Call{Inject}{V[i],\mathcal C_i,g_i}$
  \State Save execution state
  \If{$V$ has a next subtask $v_{\mathrm{next}}$}
    \State $r_{\mathrm{next}} \gets (q,v_{\mathrm{next}},V,\omega,O)$
    \State $\bar M \gets \Call{Freeze}{M}$
    \State $f \gets \Call{AsyncPrepare}{r_{\mathrm{next}},\bar M}$
  \Else
    \State $f \gets \varnothing$
  \EndIf
  \State $o_i \gets \Call{MainAgent}{q,v_i^+,\omega,O,\mathrm{tools}}$
  \State Commit $o_i$; update $M$ via Algorithm~\ref{alg:memory-update}
  \State Update $O$, $\omega$, and the remaining plan $V$
  \If{the task chain is complete}
    \State \textbf{break}
  \EndIf
  \State $v_{\mathrm{next}} \gets$ first uncommitted subtask
  \State $c \gets \Call{Wait}{f}$ if $f\ne\varnothing$; otherwise $\varnothing$
  \If{$c\ne\varnothing$ and its intent differs from revised $V$}
    \State $c \gets \varnothing$
  \EndIf
  \State $r_{\mathrm{next}} \gets (q,v_{\mathrm{next}},V,\omega,O)$
  \If{$c\ne\varnothing$}
    \State $D_{\mathrm{next}} \gets \Call{Prepare}{r_{\mathrm{next}},M,\mathrm{Continue},c}$
  \Else
    \State $D_{\mathrm{next}} \gets \Call{Prepare}{r_{\mathrm{next}},M,\mathrm{Prepare}}$
  \EndIf
  \State $i \gets \Call{Index}{v_{\mathrm{next}}}$; $D_i \gets D_{\mathrm{next}}$
\EndWhile
\State \Return \Call{Aggregate}{$O$}
\end{algorithmic}
\end{algorithm}

\begin{algorithm}[tbp]
\renewcommand{\thealgorithm}{B2a}
\caption{Bounded Search/Final preparation}
\label{alg:prepare}
\small
\begin{algorithmic}[1]
\Require Context $r=(q,v_i,V,\omega,O)$; memory $M$ or a snapshot; policy; phase $\phi$; optional candidate $c$; retrieval settings; search limit $s_{\max}$
\Ensure Final guidance, context, and search-round count
\State $\bar M \gets \Call{Freeze}{M}$
\State $u \gets q + \operatorname{instruction}(v_i)$
\State $S \gets \{(W,u),(P,u)\}$; $\eta \gets [\,]$
\For{$j=0,\ldots,s_{\max}$}
  \State $R \gets \Call{Retrieve}{\bar M,S}$
  \State Append $(S,R)$ to $\eta$
  \State $x \gets \Call{BuildContext}{r,R,\eta,\phi,c}$
  \State $a \gets \Call{PersonaAgent}{x}$
  \If{$a.\mathrm{action}=\mathrm{Final}$}
    \State \Return $(a.\mathrm{memory\_points},a.\mathrm{answer},x,j)$
  \EndIf
  \If{$a.\mathrm{action}\ne\mathrm{Search}$ or $j=s_{\max}$}
    \State Stop preparation
  \EndIf
  \State $S \gets S\cup a.\mathrm{searches}$
\EndFor
\end{algorithmic}
\end{algorithm}
Workspace state corresponds to $\omega_i$ and retrieval history to $\eta$. Continue receives the latest state, completed results, and updated memories and returns fresh conditions/guidance for the next subtask.

\FloatBarrier
\subsection{GRPO Objective}
\label{app:grpo}
\setcounter{equation}{18}
For candidate reward $R_j$, group-normalized advantages are
\begin{equation}
\begin{aligned}
\hat A_j&=\frac{R_j-\mu_R}{\sigma_R+\varepsilon_A},\\
\mu_R&=\frac1G\sum_{j=1}^{G}R_j,\\
\sigma_R&=\sqrt{\frac1G\sum_{j=1}^{G}(R_j-\mu_R)^2}.
\end{aligned}
\label{eq:19}
\end{equation}
For generated action token $a_{j,t}$, let $z_{j,t}$ contain its preceding context, generated actions, and retrieval returns. The importance ratio and clipping are
\begin{equation}
\begin{aligned}
\rho_{j,t}(\theta)&=\frac{\pi_\theta(a_{j,t}\mid z_{j,t})}{\pi_{\mathrm{old}}(a_{j,t}\mid z_{j,t})},\\
\bar\rho_{j,t}&=\operatorname{clip}\left(\rho_{j,t},1-\epsilon_{\mathrm{clip}},1+\epsilon_{\mathrm{clip}}\right).
\end{aligned}
\label{eq:20}
\end{equation}
With generated-token positions $\mathcal I_j$, the update maximizes
\begin{equation}
\begin{aligned}
J_{\mathrm{RL}}(\theta)
={}&\mathbb E\Bigg[\frac1G\sum_{j=1}^{G}\frac1{|\mathcal I_j|}
\sum_{t\in\mathcal I_j}\\
&\Big(\min(\rho_{j,t}\hat A_j,\bar\rho_{j,t}\hat A_j)\\
&\qquad-\beta D_{\mathrm{KL},j,t}\Big)\Bigg].
\end{aligned}
\label{eq:21}
\end{equation}
Here $D_{\mathrm{KL},j,t}$ is the divergence from $\pi_\theta(\cdot\mid z_{j,t})$ to the fixed SFT reference $\pi_{\mathrm{ref}}(\cdot\mid z_{j,t})$, and $\pi_{\mathrm{old}}$ generates candidate sequences. $\varepsilon_A$ provides numerical stability, $\epsilon_{\mathrm{clip}}$ is the clipping threshold, and $\beta$ controls KL regularization. Tool returns form context for later decisions; loss applies to PersonaAgent outputs.

\begin{algorithm}[tbp]
\renewcommand{\thealgorithm}{B3}
\caption{Policy optimization}
\label{alg:policy-optimization}
\small
\begin{algorithmic}[1]
\Require SFT policy; training inputs; group size $G$
\Ensure Trained PersonaAgent policy $\pi_\theta$
\State Initialize $\pi_\theta$ from SFT; freeze $\pi_{\mathrm{ref}}$
\ForAll{training batches}
  \ForAll{inputs $x$ in the batch}
    \State Generate $G$ Search/Final sequences with $\pi_\theta$
    \State Score final conditions using Equation~\ref{eq:18}
    \State Compute advantages using Equation~\ref{eq:19}
    \State Compute clipped ratios using Equation~\ref{eq:20}
    \State Update $\pi_\theta$ with Equation~\ref{eq:21}
  \EndFor
\EndFor
\State \Return $\pi_\theta$
\end{algorithmic}
\end{algorithm}

\FloatBarrier
\section{Prompt Templates}
\label{app:prompts}
\raggedbottom
\tcbset{paperprompt/.append style={top=4pt,bottom=4pt,before skip=6pt,after skip=6pt}}
The templates follow the two memory types, subtask-level guidance, and training process. Inputs are structured JSON; double-braced variables are filled at call time. Workspace \code{env}/\code{task} map to $M_{\mathrm{env}}$/$M_{\mathrm{task}}$, and Persona \code{topic}/\code{preference}/\code{strategy} map to $k_l$/$v_l$/$\sigma_l$. The English system prompts below are retained from the aligned source appendix.

\subsection{Structured Memory Extraction}
Initial histories and online subtask outcomes use the same prompt. One call implements $F_W$ and $F_P$. The input record's request, interactions/observations, artifacts, and feedback/outcome correspond to $(q_j,\xi_j,A_j,f_j)$. The model extracts content; storage records IDs, sources, update times, and user/workspace scope.
\promptfile{C.1 / System prompt}{prompts/c1_1.txt}
\promptfile{C.1 / Input}{prompts/c1_2.txt}
Environment information enters \code{env}; goals, constraints, progress, result summaries, and artifacts enter \code{task}. Interaction summaries populate the relevant environment or previous-result fields, and reusable preferences enter \code{persona}.

\subsection{Ordered Subtask Planning}
The main agent plans ordered subtasks, each potentially containing multiple tool operations. JSON \code{steps} represents $\mathcal T$ and each item represents $s_i$; names link results to guidance.
\promptfile{C.2 / System prompt}{prompts/c2_1.txt}
\promptfile{C.2 / Input}{prompts/c2_2.txt}
The agent revises unexecuted subtasks using actual outcomes and prepares guidance for the new intent.

\subsection{Guidance Preparation and Retrieval}
Inputs correspond to $x_i=(q,s_i,\omega_i,O_{<i})$ and $R_i$. The request, subtask, workspace state, and previous outputs supply execution context. Memories provide retrieved Workspace/Persona items. Retrieval history represents $\eta$, and candidate guidance represents $\hat D$.
\promptfile{C.3 / System prompt}{prompts/c3_1.txt}
\promptfile{C.3 / Decision context}{prompts/c3_2.txt}
\code{memory\_points} corresponds to $\mathcal C_i$ and \code{answer} to $g_i$. Search adds queries and results to retrieval history and invokes the same policy again; Final ends preparation. Retrieved content follows the shared memory schema.

\subsection{Injection Before Subtask Execution}
Conditions and guidance are added before execution, implementing $s_i^+=\operatorname{Inject}(s_i,\mathcal C_i,g_i)$. The executor uses benchmark task tools and interaction protocols.
\promptfile{C.4 / Execution supplement}{prompts/c4_1.txt}
\promptfile{C.4 / Injected context}{prompts/c4_2.txt}
The completed subtask's interactions, observations, artifacts, and feedback become an extraction record for the subsequent memory update.

\subsection{Training Annotations}
Scripts synthesize profiles, behavior histories, and web tasks following Persona2Web. GPT-5.2-Codex annotates conditions and action demonstrations; policy contexts use the same preparation/updating schemas.
\subsubsection{Reference Conditions}
Reference conditions $\mathcal G=\{(k_j,v_j)\}$ are annotated using the request, workspace state, historical memory, and completed results.
\promptfile{C.5 / Reference-conditions prompt}{prompts/c5_1.txt}
\promptfile{C.5 / Reference-conditions input}{prompts/c5_2.txt}
Keys and values form the scoring reference; evidence IDs record sources. Both memory types use the extraction schema.
\subsubsection{Actions and Guidance}
Action demonstrations are annotated using the guidance-preparation prompt with the additional instruction and input below. Updating examples include candidate guidance and the latest results in the decision context.
\promptfile{C.5 / Annotation instruction}{prompts/c5_3.txt}
\promptfile{C.5 / Annotation input}{prompts/c5_4.txt}
Reference conditions are annotation targets. The context includes subtask state, retrieval returns, and history. After each annotated Search, the system executes retrieval and supplies the updated context for the next action annotation. Initial preparation has no candidate; update demonstrations include the candidate, latest state, and completed results. Annotated action sequences and guidance train SFT, while reference conditions score RL.

\end{document}